\documentclass[a4paper,conference]{IEEEtran}
\ifCLASSINFOpdf
  \usepackage[pdftex]{graphicx}
  \graphicspath{
  {./figures/bn_structures/4nodes/}
  {./figures/ste_example/}
  {./figures/tradeoff_nparameters_error/letter/}
  {./figures/tradeoff_nparameters_error/mnist/}
  {./figures/tradeoff_nparameters_error/satimage/}
  {./figures/tradeoff_nparameters_error/usps/}
  {./figures/compare_nn_bnc_equal_memory/letter/}
  {./figures/compare_nn_bnc_equal_memory/mnist/}
  {./figures/compare_nn_bnc_equal_memory/satimage/}
  {./figures/compare_nn_bnc_equal_memory/usps/}
  {./figures/compare_nn_bnc_mem_ops_err/letter/}
  {./figures/compare_nn_bnc_mem_ops_err/mnist/}
  {./figures/compare_nn_bnc_mem_ops_err/satimage/}
  {./figures/compare_nn_bnc_mem_ops_err/usps/}
  {./figures/compare_cnn_bnc_ops/mnist/}
  {./figures/compare_cnn_bnc_ops/usps/}
  {./figures/compare_bnc_nb_tan/letter/}
  {./figures/compare_bnc_nb_tan/mnist/}
  {./figures/compare_bnc_nb_tan/satimage/}
  {./figures/compare_bnc_nb_tan/usps/}
  }
\else
\fi
\ifCLASSOPTIONcompsoc
  \usepackage[caption=false,font=normalsize,labelfont=sf,textfont=sf]{subfig}
\else
  \usepackage[caption=false,font=footnotesize]{subfig}
\fi
\usepackage{xcolor}
\usepackage{amsmath}
\usepackage{amssymb}
\newcommand{\argmax}{\operatornamewithlimits{argmax}}
\newcommand{\matW}{\mathbf{W}}
\newcommand{\graph}{\mathcal{G}}
\newcommand{\rvxset}{\boldsymbol{X}}
\newcommand{\rvc}{\ensuremath{C}}
\newcommand{\rvx}{\ensuremath{X}}

\newcommand{\parent}{\mathrm{pa}}
\newcommand{\clip}{\mbox{clip}}
\newcommand{\round}{\mbox{round}}
\DeclareMathOperator{\sign}{sign}
\DeclareMathOperator{\Unif}{\mathcal{U}}

\newcommand{\myvert}{\hspace{1pt} \vert \hspace{1pt}}
\newcommand{\loss}{\mathcal{L}}
\newcommand{\veca}{\mathbf{a}}
\newcommand{\vecx}{\mathbf{x}}
\newcommand{\vecb}{\mathbf{b}}
\newcommand{\vecs}{\mathbf{s}}
\newcommand{\vecl}{\boldsymbol{l}}
\newcommand{\vecbeta}{\boldsymbol{\beta}}
\newcommand{\vecgamma}{\boldsymbol{\gamma}}

\newcommand{\vecPhi}{\Phi}
\newcommand{\vectheta}{\boldsymbol{\theta}}
\newcommand{\vecTheta}{\boldsymbol{\Theta}}
\newcommand{\vecrho}{\boldsymbol{\rho}}

\newcommand{\E}{\mathbb{E}}
\newcommand{\indicator}[1][]{{\mathbb{I} \def\temp{#1} \ifx\temp\empty \else \left[ #1 \right] \fi}} 
\newcommand{\structuretradeoff}{\lambda_{\mathrm{MS}}}

\usepackage{hyperref}

\begin{document}
%
\title{On Resource-Efficient Bayesian Network Classifiers and Deep Neural Networks}


\author{\IEEEauthorblockN{Wolfgang Roth and\\Franz Pernkopf}
\IEEEauthorblockA{Signal Processing and Speech Communication Laboratory,\\
Graz University of Technology, Graz, Austria\\
Email: \{roth,pernkopf\}@tugraz.at}
\and
\IEEEauthorblockN{G\"unther Schindler and\\Holger Fr\"oning}
\IEEEauthorblockA{Institute of Computer Engineering,\\Ruprecht Karls University, Heidelberg, Germany\\
Email: \{guenther.schindler,holger.froening\}@ziti.uni-heidelberg.de}}


%


\maketitle

\begin{abstract}
We present two methods to reduce the complexity of Bayesian network (BN) classifiers.
First, we introduce quantization-aware training using the straight-through gradient estimator to quantize the parameters of BNs to few bits.
Second, we extend a recently proposed differentiable tree-augmented na\"ive Bayes (TAN) structure learning approach by also considering the model size.
Both methods are motivated by recent developments in the deep learning community, and they provide effective means to trade off between model size and prediction accuracy, which is demonstrated in extensive experiments.
Furthermore, we contrast quantized BN classifiers with quantized deep neural networks (DNNs) for small-scale scenarios which have hardly been investigated in the literature.
We show Pareto optimal models with respect to model size, number of operations, and test error and find that both model classes are viable options.
\end{abstract}


%
\IEEEpeerreviewmaketitle

\section{Introduction} \label{sec:introduction}
One key factor for the tremendous successes of deep learning in a wide variety of applications are the ever growing sizes of network architectures and the availability of dedicated massively parallel hardware such as GPUs and TPUs.
As a result, many interesting applications do not benefit from the capabilities of deep learning since deep neural networks (DNNs) are often too large and cannot be deployed on resource-constrained devices with limited memory, computation power, and battery capacity.
This has sparked the development of a variety of methods for reducing the complexity of DNNs.
These methods are as diverse as weight pruning, quantization, exploiting structural properties that allow for resource-efficient computation, and, very recently, neural architecture search to discover efficient architectures automatically \cite{Roth2020a}.
However, most of the literature considers medium to large-scale datasets that require a moderate architecture size to achieve a decent accuracy.
Consequently, also the resulting DNNs after applying the respective methods are still too large for resource-constrained devices.

In this paper, we transfer quantization techniques from the recent DNN literature to an inherently resource-efficient model class, namely Bayesian Network (BN) classifiers with na\"ive Bayes or tree-augmented na\"ive Bayes (TAN) structure.
For datasets with $C$ classes and $D$ \emph{discrete} input features, a BN classifier efficiently computes predictions by accumulating $(D+1) \cdot C$ log-probabilities and determining the most probable class.
Notably, no other operations, such as multiplications, are required.

In particular, we employ quantization-aware training using the straight-through gradient estimator (STE) \cite{Bengio2013}.
The STE is used to approximate the zero derivative of piecewise constant functions $f$, such as a quantizer, by the non-zero derivative of a similar function $\tilde{f}$.
This allows us to incorporate quantizers and other piecewise constant functions in a computation graph and to perform gradient-based learning using backpropagation.

During training, our quantization method maintains a set of real-valued auxiliary parameters $\vectheta$ that are quantized to few bits during forward propagation to obtain $\vectheta_q$.
During backpropagation, the gradient is computed with respect to the quantized parameters $\vectheta_q$ which is then passed ``straight-through'' to update the real-valued parameters $\vectheta$.
This procedure is typically more effective than performing quantization as a post-processing step, since the real-valued parameters $\vectheta$ become robust to quantization during training.
After training, the model is deployed using the quantized parameters $\vectheta_q$.
This paradigm has been extensively used for quantization in the deep learning literature \cite{Hubara2016,Zhou2016,Rastegari2016}.

Furthermore, we extend the recently proposed differentiable TAN structure learning approach from \cite{Roth2020b} by also taking the model size into account.
Their approach introduces a distribution over TAN structures which is jointly trained with the BN parameters using gradient-based learning.
After training, the most probable TAN structure is selected.
In this paper, we introduce an additional loss term that penalizes the number of parameters of a specific TAN structure, which allows us to effectively trade off between accuracy and model size.

Notably, this method is also inspired by differentiable structure learning approaches from the deep learning community \cite{Liu2019,Cai2019}.
Typical BN structure learning algorithms rely on greedy hill-climbing heuristics for combinatorial optimization and are not suitable for gradient-based optimization \cite{Grossman2004,Teyssier2005,Pernkopf2013}.
Note that our method bears resemblance to the minimum description length principle where one also aims to optimize for model fit and model size \cite{Friedman1997}.

In extensive experiments, we contrast quantized BN classifiers with small-scale quantized DNNs with respect to (i) model size, (ii) number of operations required for predictions, and (iii) the prediction accuracy.
We investigate Pareto optimal models with respect to these three dimensions and find that no model class can be excluded a priori.
Furthermore, we compare our quantization method with a specialized branch-and-bound approach that directly operates on the discrete parameter space of BNs \cite{Tschiatschek2014}.
Our quantization method does not only achieve higher accuracy, but it also takes much less training time than the computationally intensive branch-and-bound algorithm.
We demonstrate that our structure learning approach can be used to generate a trajectory of Pareto optimal BN classifiers with respect to accuracy and model size.

In summary, the main contributions of this paper are:
\begin{itemize}
 \item Quantization-aware learning of BN classifiers
 \item Differentiable model-size-aware TAN structure learning
 \item A comprehensive comparison of quantized DNNs and BN classifiers
\end{itemize}
Code is available online at \url{https://github.com/wroth8/bnc}

\section{Background} \label{sec:background}

\subsection{Bayesian Network Classifiers} \label{sec:bayesian_network_classifiers}
We denote random variables as uppercase letters $\rvx$ and $\rvc$ and instantiations of these random variables as lowercase letters $x$ and $c$.
Let $\rvxset = \{\rvx_1, \ldots, \rvx_D\}$ be a multivariate random variable.
A Bayesian Network (BN) is a graphical representation of a probability distribution $p(\rvxset)$ as a directed acyclic graph $\graph$ whose nodes correspond to the random variables $\rvx_i$.
More specifically, the graph $\graph$ determines a factorization of $p(\rvxset)$ according to
\begin{align}
  p(\rvxset) = \prod_{i=1}^{D} p\left(\rvx_i \myvert \parent(\rvx_i)\right), \label{eq:bn_factorization}
\end{align}
where $\parent(\rvx_i)$ is the set of parents of $\rvx_i$ in $\graph$.
This factorization allows us to specify the full joint distribution $p(\rvxset)$ by the individual factors $p(\rvx_i \myvert \parent(\rvx_i))$.
We consider distributions over \emph{discrete} random variables such that each conditional distribution $p(\rvx_i \myvert \parent(\rvx_i))$ can be represented as a conditional probability table (CPT) $\vectheta_{i|\parent(i)}$.
The joint distribution $p(\rvxset)$ is then specified by the collection of CPTs $\vectheta = \{\vectheta_{1|\parent(1)}, \ldots, \vectheta_{D|\parent(D)} \}$ of all random variables $\rvxset$.

When considering the task of classification in particular, we are given an additional class random variable $\rvc$ and assume that a BN is used to model the joint distribution $p(\rvxset, \rvc)$.
In this context, the variables $\rvxset$ are called input features.
We can then construct a probabilistic classifier by assigning an input $\vecx$ to the conditionally most probable class $c$ according to
\begin{align}
  \argmax_{c} p(c \myvert \vecx) = \argmax_{c} p(\vecx, c) = \argmax_{c} \log p(\vecx, c). \label{eq:probabilistic_classifier}
\end{align}
Assuming that the individual factors of \eqref{eq:bn_factorization} can be computed efficiently, classification according to \eqref{eq:probabilistic_classifier} is particularly convenient by accumulating only $D+1$ log-probabilities
\begin{align}
 \log p(\rvxset,\rvc) = \log p\left(\rvc \myvert \parent(\rvc) \right) + \sum_{i=1}^{D} \log p\left(\rvx_i \myvert \parent(\rvx_i)\right) \label{eq:bn_log_probabilities}
\end{align}
for each class $c$ and reporting the most probable class.

However, the situation becomes problematic concerning the size of the CPTs $\vectheta_{i|\parent(i)}$ which is determined by the number of values that $\rvx_i$ and $\parent(\rvx_i)$ can take jointly.
Consequently, assuming that each random variable $\rvx_i$ can take at least two values, the size of $\rvx_i$'s CPT grows exponentially with the number of parents $|\parent(\rvx_i)|$, which can become a computational bottleneck even for few parents.
Therefore, it is desirable to maintain graph structures where each node has few parents such that inference tasks remain feasible.
In this paper, we consider two commonly used types of structures for BN classifiers which restrict the number of conditioning parents, namely the na\"ive Bayes structure and TAN structures.

\begin{figure}[!t]
\centering
\subfloat[Na\"ive Bayes structure]{\includegraphics[width=0.2\textwidth]{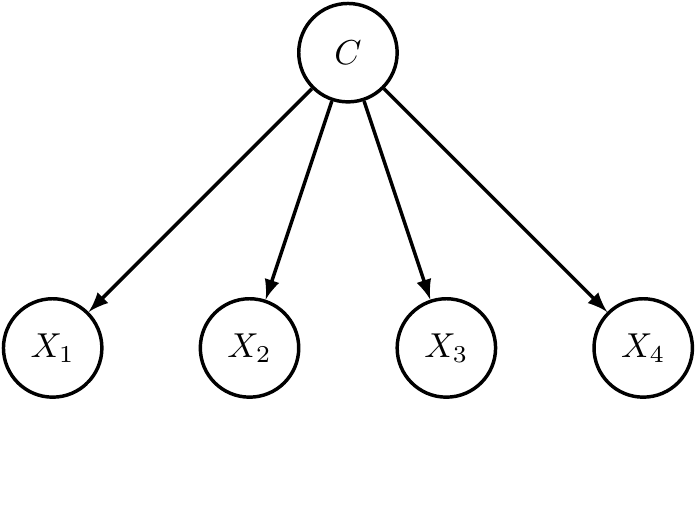}\label{fig:naive_bayes}}
\hfil
\subfloat[TAN structure]{\includegraphics[width=0.2\textwidth]{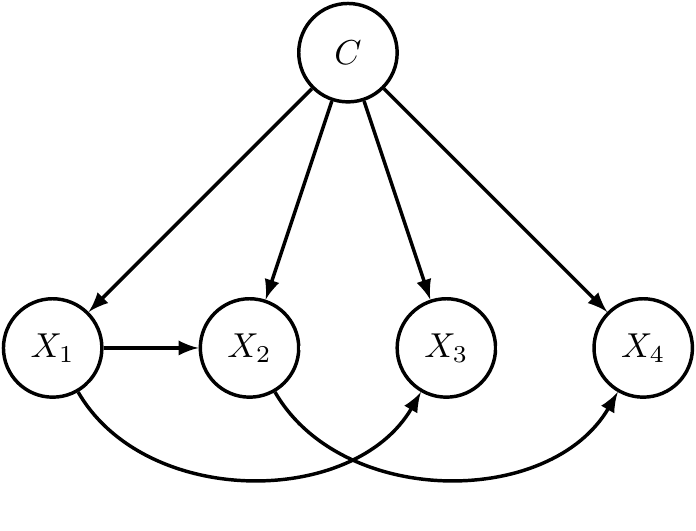}\label{fig:tan_structure}}
\caption{%
\protect\subref{fig:naive_bayes} Graphical representation of the na\"ive Bayes model as a BN.
The na\"ive Bayes model assumes conditional independence of the inputs $\rvxset$ given the class $\rvc$.
\protect\subref{fig:tan_structure} Example TAN structure.
The TAN structure allows each input $\rvx_i$, in addition to the class variable $\rvc$, to depend on a single other input $\rvx_j$.
}
\label{fig:bayesian_networks}
\end{figure}

\subsubsection{Na\"ive Bayes Structure} \label{sec:naive_bayes_structure}
The na\"ive Bayes structure is illustrated in \figurename~\ref{fig:naive_bayes}.
The graph $\graph$ contains a single root node $\rvc$ which is the sole parent of each feature node $\rvx_i$.
The factorization induced by the na\"ive Bayes assumption is given by
\begin{align}
  p(\rvxset, \rvc) = p(\rvc) \prod_{i=1}^{D} p\left(\rvx_i \myvert \rvc \right). \label{eq:joint_probability_nb}
\end{align}
The na\"ive Bayes model assumes that all inputs $\rvxset$ are conditionally independent given the class $\rvc$.
Although this independence assumption rarely holds in practice, na\"ive Bayes models often perform reasonably well while requiring only few parameters and allowing for fast inference.

\subsubsection{Tree-augmented Na\"ive Bayes (TAN) Structure} \label{sec:tan_structure}
The TAN structure generalizes the na\"ive Bayes structure by allowing each feature $\rvx_i$ --- in addition to the class variable $\rvc$ --- to directly depend on at most one other feature $\rvx_j$.
An example TAN structure is illustrated in \figurename~\ref{fig:tan_structure}.
The factorization of a TAN BN is given by
\begin{align}
  p(\rvxset, \rvc) = p(\rvc) \prod_{i=1}^{D} p\left(\rvx_i \myvert \parent(\rvx_i) \right), \label{eq:joint_probability_tan}
\end{align}
subject to $|\parent(\rvx_i)| \leq 2$ and $\rvc \in \parent(\rvx_i)$.
As we will see in Section~\ref{sec:experiments}, this relaxation of the graph structure typically improves the predictive performance, but it also introduces a substantial memory overhead due to larger CPTs.

As opposed to the na\"ive Bayes model, the TAN structure is not fixed, and we can utilize this freedom to perform structure learning in order to balance accuracy and model size.
However, this is not straightforward as the number of possible TAN structures is exponential in the number of input features $D$.
In Section~\ref{sec:structure_learning}, we present a differentiable method for jointly training the graph structure $\graph$ and the CPTs $\vectheta$ that favors smaller models.

\subsection{Deep Neural Networks} \label{sec:background_dnn}
DNNs are a class of discriminative models that directly model the conditional probability $p(\rvc \myvert \rvxset)$.
A DNN is organized in $L$ layers, where each layer computes an affine transformation followed by an activation function $h$, i.e.,
\begin{align}
  \veca^l &= \matW^l \otimes \vecx^{l-1} + \vecb^l \label{eq:nn_linear} \\
  \vecx^l &= h^l(\veca^l), \label{eq:nn_non_linear}
\end{align}
where $\otimes$ refers to either a matrix-vector multiplication (fully connected layer) or a linear convolution.
Here, $\vecx^0$ corresponds to the given input features, and $\vecx^L$ is the output of the DNN.
A DNN that performs at least one convolution in \eqref{eq:nn_linear} is called a convolutional neural network (CNN).
The parameters $\vectheta$ of a DNN are given by the weight tensors $\{\matW^1,\ldots,\matW^L\}$, the bias vectors $\{\vecb^1,\ldots,\vecb^L\}$, and the batch normalization (see below) parameters $\{\vecbeta^1, \ldots, \vecbeta^L\}$ and $\{\vecgamma^1, \ldots, \vecgamma^L\}$ of all layers.

For $l < L$, we consider the element-wise non-linear ReLU activation function $h^l(a) = \max(0, a)$ and the element-wise sign activation function $h^l(a) = \indicator[x \geq 0] - \indicator[x < 0]$, where $\indicator$ denotes the indicator function.
In the output layer $l = L$, we apply the softmax function $h_i^L(\veca^L) = \exp(a_i^L) / \sum_{j} \exp(a_j^L)$ which transforms the output activations $\veca^L$ into probabilities $\vecx^L$ that we interpret as conditional class probabilities $p(c \myvert \vecx^0)$.

Many modern DNN architectures perform \emph{batch normalization} \cite{Ioffe2015} after the affine transformation \eqref{eq:nn_linear} during learning.
In each layer, batch normalization transforms the individual activations $a_i^l$ to have zero mean and unit variance across all data samples, which are then subject to a feature-wise (or, in the case of convolutions, channel-wise) affine transformation with the learnable batch normalization parameters $\vecbeta^l$ and $\vecgamma^l$.

\subsection{Similarities between BN Classifiers and DNNs} \label{sec:similarities_bn_dnns}
The output layer of a DNN performs the same computations as a linear logistic regression model, i.e., it performs an affine transformation $\matW^{L} \vecx^{L-1} + \vecb^{L}$ and reports the most probable class.
This is similar to the computations performed by a BN classifier.
Indeed, for a na\"ive Bayes model, by encoding all discrete input features $x_i$ as one-hot vectors $\bar{\vecx}_i$ which are stacked into a single sparse vector $\bar{\vecx}$, we can cast the computation of $\log p(\rvxset, \rvc)$ as an affine transformation $\bar{\matW} \bar{\vecx} + \bar{\vecb}$, where $\bar{\matW}$ contains entries from the CPTs $\vectheta$, and $\bar{\vecb}$ corresponds to $\log p(C)$.
For TAN structures, a similar transformation can be obtained by a one-hot encoding of the values that $\rvx_i$ and its additional parent $X_j$ take jointly.

This suggests that well-established methods for DNNs might be transferable to BN classifiers.
In the next section, we propose quantization for BN parameters based on methods that are successfully applied in the deep learning community.

\section{Quantization-aware Training} \label{sec:quantization_aware_training}

\subsection{Straight-through Gradient Estimator (STE)} \label{sec:ste}
Learning models that employ piecewise constant functions, such as quantizers or the sign function, is problematic as their derivatives are zero almost everywhere.
During backpropagation, these functions prevent the gradient to flow backwards, such that gradient-based learning cannot be performed.
In recent years, the STE as introduced in \cite{Bengio2013} has become a widely used tool to perform backpropagation through zero-gradient or non-differentiable building blocks in computation graphs.

Let $f$ be a function whose derivative does not exist or is zero almost everywhere.
Furthermore, let $u = f(v)$ be a value of the computation graph that is computed during forward propagation, i.e., when evaluating some loss function $\loss$.
Then, the STE approximates the partial derivative of $\loss$ with respect to $v$ during backpropagation as
\begin{align}
  \frac{\partial \loss}{\partial v} = \frac{\partial \loss}{\partial u} \cdot \frac{\partial f}{\partial v} \approx \frac{\partial \loss}{\partial u} \cdot \tilde{f}^{\prime}(v), \label{eq:ste}
\end{align}
where $\tilde{f}(v) \approx f(v)$ and $\tilde{f}^{\prime}(v)$ is non-zero.\footnote{For simplicity, we have assumed that $u$ is the only node in the computation graph depending on $v$ --- otherwise \eqref{eq:ste} would consist of more terms.}
This allows gra\-di\-ents to flow through $f$ such that parameters can still be updated using gradient-based learning.

\begin{figure}[!t]
\centering
\includegraphics[width=\columnwidth]{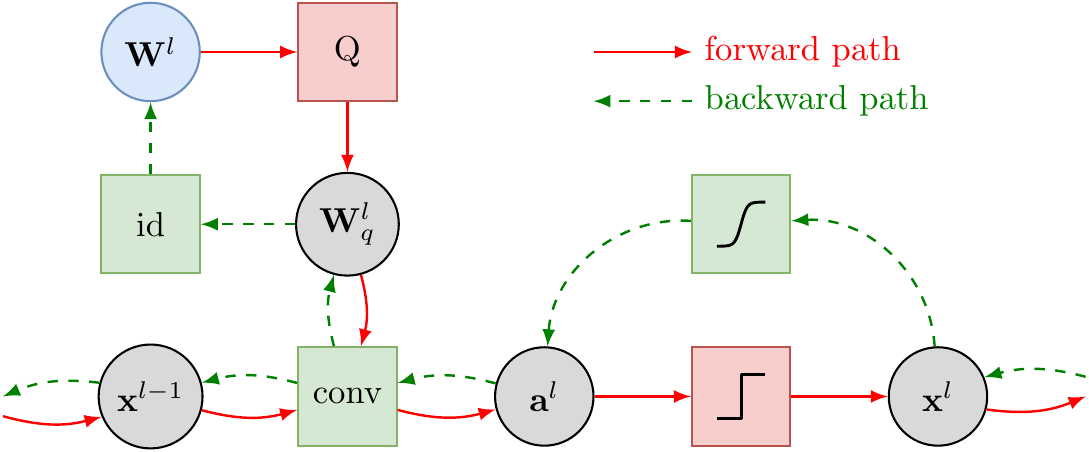}
\caption{%
Straight-through gradient estimation (STE) in a convolutional layer.
The green boxes indicate differentiable functions.
The red boxes indicate piecewise constant functions whose gradient is zero almost everywhere.
The blue box indicates learnable weights.
$Q$ denotes a quantization function, e.g., a rounding function, and \emph{id} denotes the identity function.
During forward propagation, the red path is followed, whereas during backpropagation, the dashed green path is followed to avoid the red zero-gradient boxes.}
\label{fig:ste_example}
\end{figure}

The STE has been heavily used for weight and activation quantization in DNNs, which is often referred to as quantization-aware learning.
A typical quantized DNN layer is depicted in \figurename~\ref{fig:ste_example}.
During forward propagation, the real-valued weights $\matW^l$ are quantized to obtain $\matW_q^l$, and the activations $\veca^{l}$ pass through a piecewise constant activation function such as $\sign$.
During backpropagation, the real-valued weights are updated using the STE by avoiding the zero-gradient functions.
At test time, the real-valued weights $\matW^l$ are discarded and predictions are computed using the quantized weights $\matW_q^l$.

\subsection{Quantization-aware BNs} \label{sec:quantization_aware_bns}
As briefly discussed in Section~\ref{sec:bayesian_network_classifiers}, it is convenient to store the CPT parameters of BNs as log-probabilities $\vectheta$.
During training, we store the parameters as \emph{unnormalized} log-probabilities $\vecrho$.
At forward propagation, we compute the non-positive \emph{normalized} log-probabilities $\vectheta$ as
\begin{align}
  \theta_{i}^{k,\vecl} = \rho_{i}^{k,\vecl} - \log \sum_{k^{\prime}} \exp \rho_{i}^{k^{\prime},\vecl},
\end{align}
where $\theta_{i}^{k,\vecl} = \log p\left(\rvx_i = k \myvert \parent(\rvx_i) = \vecl \right)$.
Note that the normalization is necessary as otherwise the log-likelihood could be made arbitrarily large.

In \cite{Tschiatschek2014}, it is proposed to represent the normalized log-probabilities $\vectheta$ as
\begin{align}
  \theta_q = -\sum_{k = -B_F}^{B_I - 1} b_k \cdot 2^k \label{eq:bn_logit_representation},
\end{align}
where $\vecb \in \{0,1\}^{B_I+B_F}$ is a bit-mask, $B_F$ denotes the number of fractional bits, and $B_I$ denotes the number of integer bits.
To quantize $\theta \leq 0$ to the set of possible values representable by \eqref{eq:bn_logit_representation}, we apply the quantizer
\begin{align}
  q_{\mathrm{BN}}(\theta) = \clip(\round(\theta \cdot 2^{B_F}) \cdot 2^{-B_F}, \ -U, \ 0), \label{eq:quantize_bn}
\end{align}
where $\clip(v,l,u)$ is the clipping function $\min(\max(v, l), u)$ and $U = 2^{B_I} - 2^{-B_F}$ is the largest magnitude representable by \eqref{eq:bn_logit_representation}.
During backpropagation, we apply the derivative of the identity function for the STE.
Note that after quantization the log-probabilities are in general not normalized anymore.

\subsection{Quantization-aware DNNs} \label{sec:quantization_aware_dnns}
For DNNs, we quantize the weights according to
\begin{align}
  q_{\mathrm{DNN}}(w) = q\left( \frac{\clip(w, \ -1, \ 1) + 1}{2}; B \right) \cdot 2 - 1, \label{eq:quantize_dnn}
\end{align}
where $q(v; B)$ is the quantization scheme proposed in \cite{Zhou2016} which quantizes $v \in [0, 1]$ to a $B$-bit number as
\begin{align}
  q(v; B) = \frac{1}{2^B - 1} \cdot \round((2^B - 1) \cdot v). \label{eq:quantize_dorefa}
\end{align}
Again, we employ the identity function for the STE.

In case the sign activation function is used, we use a stochastic sign function during training according to
\begin{align}
  \sign_{\mathrm{stoch}}(a) = \begin{cases} 1 \quad & \mbox{if } u \leq (1 + a) / 2 \\ -1 \quad & \mbox{otherwise} \end{cases},
\end{align}
where $u \sim \Unif([0,1])$ is drawn from a uniform distribution.
During backpropagation, we employ the derivative of the hyperbolic tangent $\tanh$ for the STE.

\section{Model-size-aware TAN Structure Learning} \label{sec:structure_learning}
In this section, we introduce a differentiable TAN structure learning approach that allows us to trade off between accuracy and model size.
Therefore, we employ the recently proposed differentiable TAN structure learning approach from \cite{Roth2020b}.

To avoid the acyclicity constraint of BNs, the approach of \cite{Roth2020b} considers a fixed variable ordering $\rvx_1, \ldots, \rvx_D$.
Let $\vecs_i = (s_{i|0}, \ldots, s_{i|i-1})$ be a one-hot encoding of $\rvx_i$'s parents such that $s_{i|j} = 1$ iff $\parent(\rvx_i) = \{ \rvx_j, \rvc\}$ and $s_{i|0} = 1$ iff $\parent(\rvx_i) = \{ C \}$ (i.e., no additional parent).
Then $\vecs = \{\vecs_1, \ldots, \vecs_D\}$ is an encoding of the TAN structure graph $\graph$.
Furthermore, let $\vecTheta_i = \{\vectheta_{i|0}, \ldots, \vectheta_{i|i-1}\}$ be the CPTs of all possible parents of $\rvx_i$, and let $\vecTheta = \{ \vecTheta_1, \ldots, \vecTheta_D\} \cup \{\vectheta_c\}$ be the collection of all possible CPTs, where $\vectheta_c$ specifies the class prior.
Then the log-likelihood $\log p(\rvxset, \rvc)$ of a TAN BN can be rephrased as
\begin{align}
  \log p_{\vectheta_c}(\rvc) + \sum_{i = 1}^{D} \sum_{j = 0}^{i - 1} s_{i|j} \log p_{\vectheta_{i|j}}(\rvx_i \myvert \rvx_j, \rvc), \label{eq:log_joint_structure}
\end{align}
where the subscripts of $p$ are to emphasize the dependency on the CPT parameters $\vecTheta$, and we define $X_0 = \emptyset$.
This allows us to jointly treat the CPTs $\vecTheta$ and the structure parameters $\vecs$ and, in principle, to optimize a loss $\loss(\vecTheta, \vecs)$ over both variables.

Since the structure parameters $\vecs$ are discrete, Roth and Pernkopf \cite{Roth2020b} introduced a probability distribution over $\vecs$ and formulate a structure learning loss $\loss_{\mathrm{SL}}$ as an expectation with respect to this distribution.
In particular, let $\vecPhi = (\vecPhi_1, \ldots, \vecPhi_D)$ with $\vecPhi_i = (\phi_{i|0}, \ldots, \phi_{i|i-1})$ be a collection of probability vectors over the one-hot vectors in $\vecs$, such that $\sum_{j=0}^{i-1} \phi_{i|j} = 1$ and $\phi_{i|j} \geq 0$.
The structure learning loss is then given by
\begin{align}
  \loss_{\mathrm{SL}}(\vecPhi, \vecTheta) = \E_{\vecs \sim p_{\vecPhi}} \left[ \loss(\vecTheta, \vecs) \right], \label{eq:structure_loss}
\end{align}
which is differentiable with respect to $\vecPhi$.
The structure learning loss \eqref{eq:structure_loss} can then be optimized with gradient-based methods.
After training, the most probable TAN structure according to $\vecPhi$ is selected.
Since \eqref{eq:structure_loss} comprises exponentially many terms, it is proposed to compute Monte Carlo gradients using the reparameterization trick \cite{Kingma2014} by means of the straight-through Gumbel softmax approximation \cite{Jang2017,Maddison2017}.

In this paper, we extend the structure learning loss \eqref{eq:structure_loss} with an additional expected model size (MS) term to obtain
\begin{align}
  \loss_{\mathrm{SL}}^{\mathrm{MS}}(\vecPhi, \vecTheta) = \loss_{\mathrm{SL}}(\vecPhi, \vecTheta) + \lambda_{\mathrm{MS}} \E_{\vecs \sim p_{\vecPhi}} \left[ \loss_{\mathrm{MS}} (\vecs)  \right], \label{eq:model_size_loss}
\end{align}
where $\loss_{\mathrm{MS}} (\vecs)$ returns the number of parameters in the CPTs for structure $\vecs$, and $\lambda_{\mathrm{MS}}>0$ is a trade-off hyperparameter.
Note that the second term in \eqref{eq:model_size_loss} is given by
\begin{align}
  \E_{\vecs \sim p_{\vecPhi}} \left[ \loss_{\mathrm{MS}} (\vecs)  \right] = |\vectheta_c| + \sum_{i=1}^{D} \sum_{j=0}^{i-1} \phi_{i|j} \cdot |\vectheta_{i|j}|,
\end{align}
where $|\vectheta|$ denotes the number of parameters of $\vectheta$.
Objective \eqref{eq:model_size_loss} allows us to achieve different trade-offs between accuracy and model size by careful selection of $\lambda_{\mathrm{MS}}$ while learning the CPTs $\vecTheta$ and the TAN structure $\graph$ simultaneously.

\section{Experiments} \label{sec:experiments}
\subsection{Datasets} \label{sec:datasets}
We conducted experiments on the following datasets, that were also used in \cite{Tschiatschek2014} and \cite{Roth2020b}.

\begin{itemize}
\item \textbf{Letter:} 20,000 samples, describing one of 26 English letters using 16 numerical features extracted from images, i.e., statistical moments and edge counts \cite{Dua2019}.
\item \textbf{Satimage:} 6,435 samples containing multi-spectral values of $3 \times 3$ pixel neighborhoods in satellite images, resulting in a total of 36 features.
The task is to classify the central pixel of these image patches to one of the categories red soil,  cotton crop, grey soil, damp grey soil, soil with vegetation stubble, mixture class (all types present), very damp grey soil.
 \item \textbf{USPS:} 11,000 grayscale images of size $16 \times 16$, showing handwritten digits from 0--9 obtained from zip codes of mail envelopes \cite{Hastie2009}.
Every pixel is treated as a feature.
\item \textbf{MNIST:} 70,000 grayscale images showing handwritten digits from 0--9 \cite{LeCun1998}.
The original images of size $28 \times 28$ are linearly downscaled to $14 \times 14$ pixels.
Every pixel is treated as a feature.
\end{itemize}

Except for \emph{satimage}, where we use 5-fold cross-validation, we split each dataset into two thirds of training samples and one third of test samples.
The features of each dataset were discretized using the approach from \cite{Fayyad1993}.
The average numbers of discrete values per feature are $9.1$, $11.5$, $3.4$, and $13.2$ for the respective datasets in the order presented above.
For the DNN experiments, we normalize the discretized features to zero mean and unit variance.

\subsection{Experimental Setup} \label{sec:experimental_setup}
All experiments were performed using the stochastic optimizer Adam \cite{Kingma2015} for 500 epochs.
We selected mini-batch sizes of 50 on \emph{satimage}, 100 on \emph{letter} and \emph{usps}, and 250 on \emph{mnist}.
Each experiment is performed using the two learning rates $\{3\cdot10^{-3}, 3\cdot10^{-2}\}$, and we report the superior result of the two runs at the end of optimization.
The learning rate is decayed exponentially after each epoch, such that it decreases by a factor of $10^{-3}$ over the training run.

BN classifiers were trained using the hybrid generative discriminative loss from \cite{Roth2020b}, i.e.,
\begin{align}
  \loss_{\mathrm{HYB}}(\vectheta) = \loss_{\mathrm{NLL}}(\vectheta) + \lambda_{\mathrm{HYB}} \cdot \loss_{\mathrm{LM}}(\vectheta),
\end{align}
which trades off between the generative negative log-like\-li\-hood loss $\loss_{\mathrm{NLL}}$ and a discriminative probabilistic large margin loss $\loss_{\mathrm{LM}}$.
Several works have reported improved results when training probabilistic models using a hybrid loss \cite{Peharz2013,Roth2018}.
The loss $\loss_{\mathrm{HYB}}$ is governed by three hyperparameters: a generative discriminative trade-off parameter $\lambda_{\mathrm{HYB}}$, a desired log-margin parameter $\gamma_{\mathrm{HYB}}$, and a smoothing parameter $\eta_{\mathrm{HYB}}$.
We refer the reader to \cite{Roth2020b} for details about these parameters.

The initial CPT parameters are drawn from $\Unif([-0.1, 0.1])$.
For parameter quantization using \eqref{eq:quantize_bn}, we evaluated the total number of bits $B_I + B_F \in \{1, \ldots, 8\}$.
We varied the number of integer bits $B_I \in \{1, \ldots, 6\}$ and report for each total number of bits the result of the best performing $B_I$.
Note that $B_F$ becomes negative for some configurations.
In these cases, not every integer value is a possible outcome after quantization.
We tuned the hyperparameters of $\loss_{\mathrm{HYB}}$ using random search by evaluating 100 random configurations according to $\log_{10} \lambda_{\mathrm{HYB}} \sim \Unif([1,3])$ and $\log_{10} \gamma_{\mathrm{HYB}} \sim \Unif([-1,2])$, and we used a fixed $\eta_{\mathrm{HYB}}=10$.
These hyperparameters are tuned individually for each experiment, i.e., each setting of the remaining hyperparameters is evaluated 100 times.

DNNs were trained using the cross-entropy loss.
The initial weights are drawn from a uniform distribution whose variance is determined according to \cite{Glorot2010}.
CNN experiments were only conducted on the image datasets \emph{usps} and \emph{mnist}.

\subsection{Fixed parameter memory budget} \label{sec:experiment_fixed_parameter_memory}
In the first experiment, we investigate the classification performance of several models with a fixed memory budget for their parameters.
We compare BN classifiers with na\"ive Bayes structure (BNC NB) to fully connected DNNs (FC NNs) and CNNs.
The target memory is selected as the number of bits required by BNC NB for a given bit width $B_I + B_F$.
We designed DNNs that require approximately the same memory.

For fully connected DNNs, we constrained the number of hidden units in each layer to be equal.
We evaluated the bit width $B \in \{1,\ldots,8\}$, the number of layers $L \in \{2,3,4,5\}$, and performed each experiment once with and once without batch normalization.
In case batch normalization is employed, we count the batch normalization parameters as 32 bits, resulting in 64 bits per hidden unit.
Batch normalization is not performed in the output layer, where we use biases that are counted as 32 bits per output.
We do not use biases in the hidden layers, even when batch normalization is not employed.
With this specification, the total number of bits only depends on the number of hidden units.
We select the number of hidden units by rounding the real-valued number that would exactly match the target memory.

We proceed similarly for CNNs, where we select the number of channels.
We consider CNNs with one or two convolutional layers, followed by a fully connected output layer.
After each convolutional layer, we downscale the image by a factor of two using max-pooling.
In case of two convolutional layers, the number of channels of the second layer is constrained to be twice the number of channels of the first layer.
Again, the batch normalization parameters (if used) incur 64 bits for each channel, and we employ 32 bit biases in the output layer.
The resulting real-valued number of channels is rounded separately for the first and the second hidden layer.

If not stated otherwise, DNNs treat the (normalized) discrete input features as real values.
We also perform experiments using one-hot encoded input features as outlined in Section~\ref{sec:similarities_bn_dnns}, such that BNs and DNNs treat the inputs equally.
Note that this increases the number of weights in the first layer for a given number of hidden units.

\begin{figure*}[!ht]
\centering
\subfloat{\quad \enskip \includegraphics[scale=0.532]{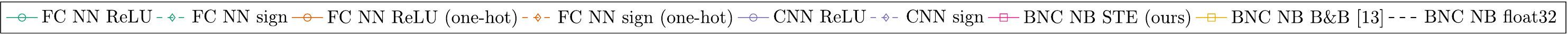}} \\[-4mm]
\setcounter{subfigure}{0}
\subfloat[letter]{\includegraphics[scale=0.53]{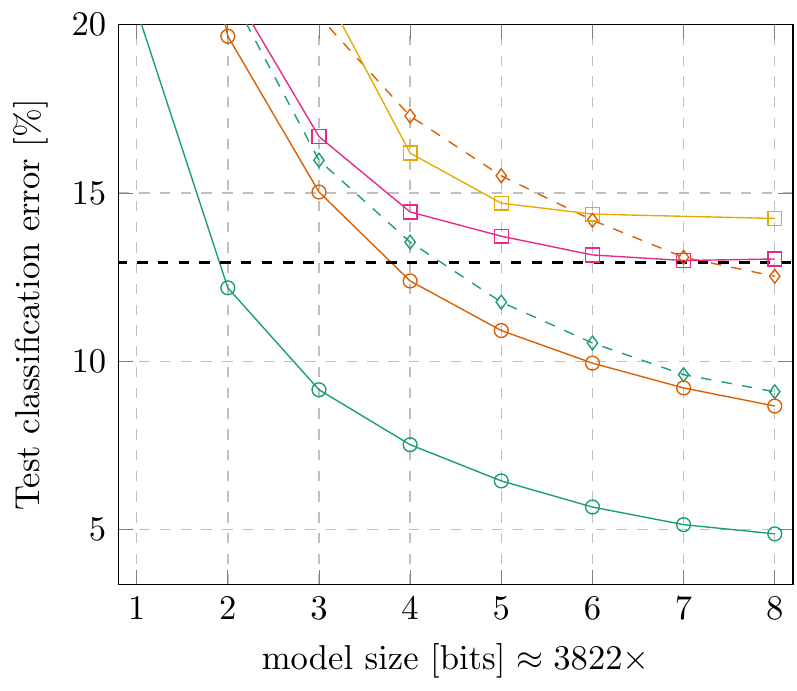}\label{fig:letter_compare_nn_bnc_equal_memory}}
\hfil
\subfloat[satimage]{\includegraphics[scale=0.53]{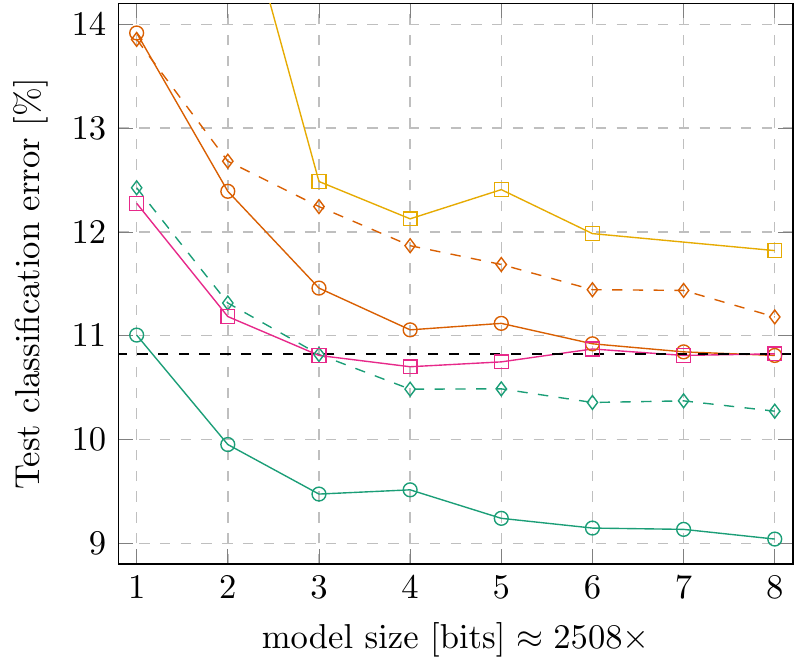}\label{fig:satimage_compare_nn_bnc_equal_memory}}
\hfil
\subfloat[usps]{\includegraphics[scale=0.53]{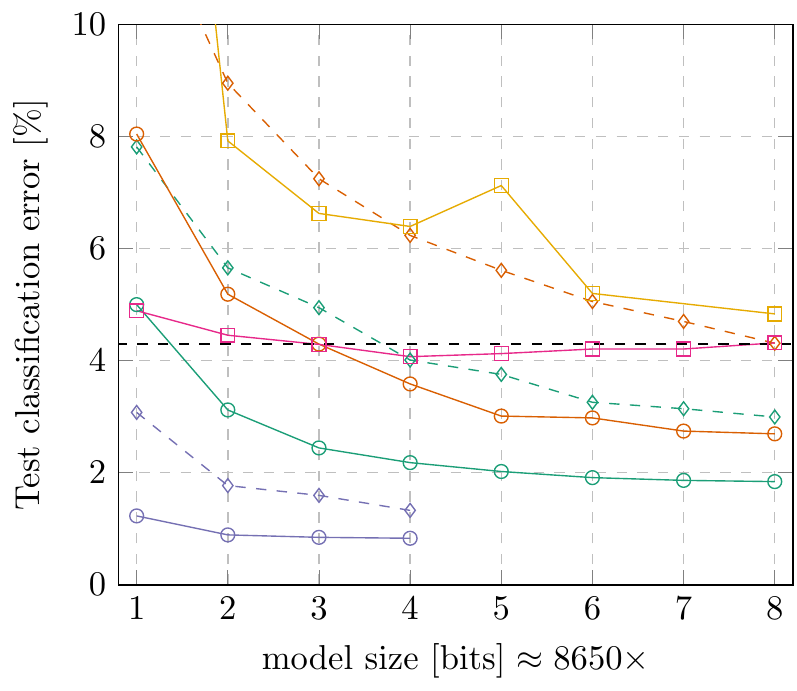}\label{fig:usps_compare_nn_bnc_equal_memory}}
\hfil
\subfloat[mnist]{\includegraphics[scale=0.53]{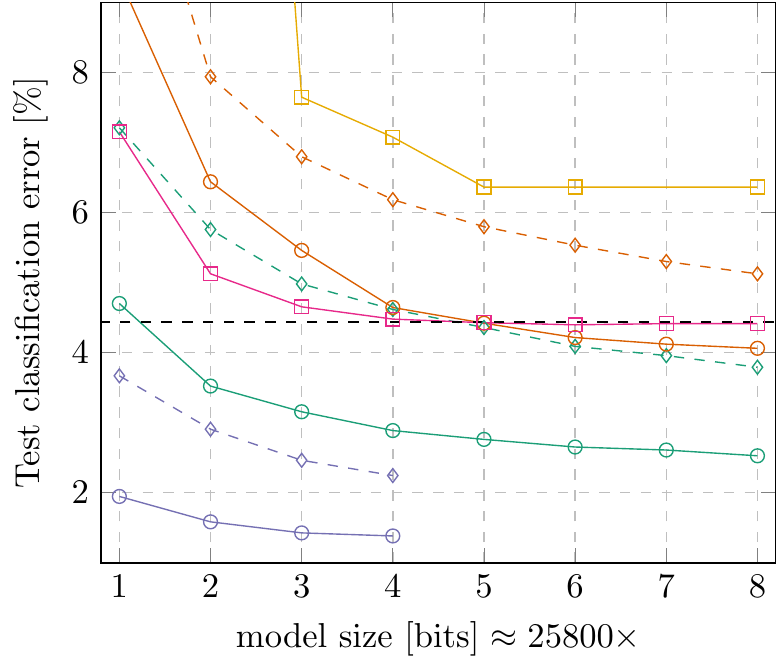}\label{fig:mnist_compare_nn_bnc_equal_memory}}
\caption{%
Test classification errors [\%] over model size budgets in bits.
The x-axis shows the model size of BN classifiers with na\"ive Bayes structure (BNC NB) for given bit widths $B_I + B_F$.
Fully connected DNNs (FC NNs) and CNNs are designed to have approximately (due to rounding) the same model size.
\vspace{-5mm}
}
\label{fig:compare_nn_bnc_equal_memory}
\end{figure*}

\begin{figure}[!ht]
\centering
\subfloat{\quad \enskip \includegraphics[scale=0.4]{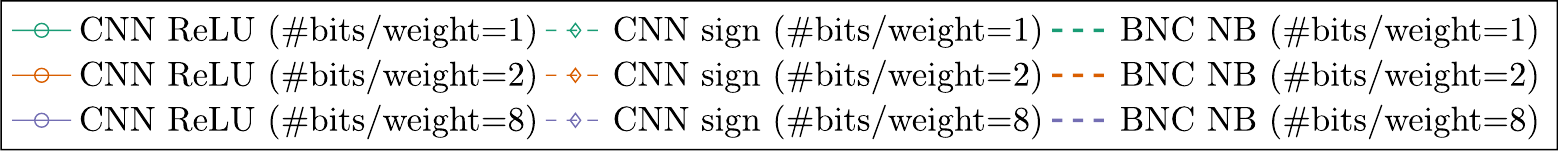}} \\[-4mm]
\setcounter{subfigure}{0}
\subfloat[usps]{\includegraphics[scale=0.53]{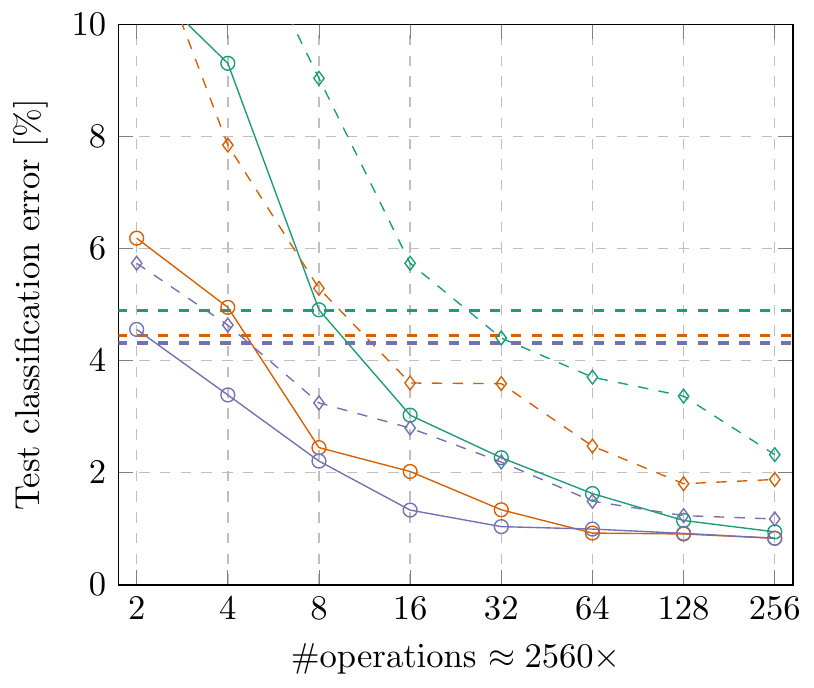}\label{fig:usps_compare_cnn_bnc_ops}}
\hfil
\subfloat[mnist]{\includegraphics[scale=0.53]{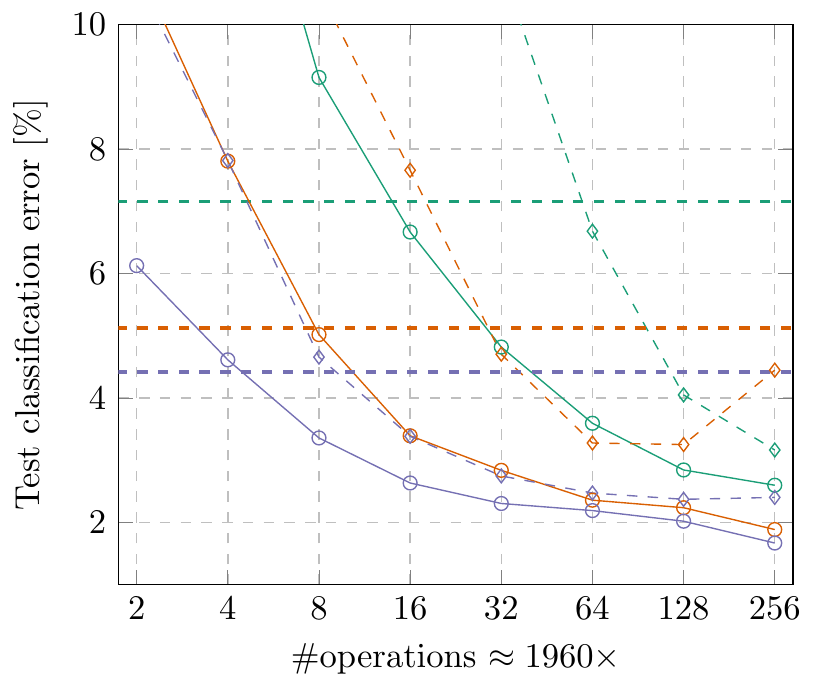}\label{fig:mnist_compare_cnn_bnc_ops}}
\caption{%
Test classification errors [\%] over fixed budgets for the number of operations.
The x-axis corresponds to multiples of the number of operations required by a BN classifier (BNC).
\vspace{-5mm}
}
\label{fig:compare_cnn_bnc_ops}
\end{figure}

The best results for a given target number of bits are shown in \figurename~\ref{fig:compare_nn_bnc_equal_memory}.
The optimal number of bits per weight is highly dataset dependent.
For instance, our BN classifiers with one bit per weight perform reasonably well on \emph{usps}, while the performance still improves up to 6--7 bits on \emph{letter}.

We confirm that CNNs are extremely memory efficient.
Even for the smallest memory budget, CNNs with ReLU activation are more accurate than all other models using the largest memory budget.
The activation function is crucial as the accuracy degrades considerably for the sign function.

Fully connected DNNs outperform BN classifiers consistently.
This is due to DNNs treating the inputs as real values, which allows them to be more memory efficient by only maintaining one weight per feature rather than one weight per feature value.
By spending the gained memory into additional layers computing intermediate representations, the performance improves.
We verified that the improvements can be attributed to the intermediate representations of DNNs, as logistic regression (one layer network) with float32 weights performs poorly on the real-valued inputs.

A fairer comparison is obtained by using one-hot encoded inputs for DNNs.
Note that for one-hot encoded inputs, it is still possible for DNNs to achieve a lower memory overhead than BN classifiers by (i) employing a hidden layer with fewer units than the number of classes and by (ii) using fewer bits per weight.
Especially the case of using fewer bits per weight highlights the importance of a DNN's capability to compute intermediate representations.
For instance, on \emph{usps}, the performance of BN classifiers with three or more bits can be obtained by a fully connected DNN using fewer bits and by spending the gained memory in an additional layer.

Our quantized BN classifier outperforms the specialized branch-and-bound method (B\&B) from \cite{Tschiatschek2014} by a large margin.
From a practical perspective, quantization-aware training fits seamlessly into existing gradient-based learning frameworks and incurs only a negligible computational overhead, whereas branch-and-bound algorithms are computationally intensive and often rely on carefully selected heuristics to reduce the runtime.
We also observed that different hyperparameters are optimal for different bit widths $B_I + B_F$.
This is in contrast to \cite{Tschiatschek2014} where the runtime of the branch-and-bound algorithm did not allow for an extensive hyperparameter search.

\subsection{Fixed number of operations budget} \label{sec:experiment_fixed_num_operations}
We compare BN classifiers with na\"ive Bayes structure (BNC NB) to CNNs with a fixed budget for the number of operations.
Since BNs require very few operations, we design CNNs that require multiples of that number of operations.
Similar to how the CNN architecture is obtained in Section~\ref{sec:experiment_fixed_parameter_memory}, we select the number of channels to match a given target number of operations.
We treat both addition and multiply-accumulate as single operations.
Batch normalization and adding biases incur one operation per hidden unit.

\figurename~\ref{fig:compare_cnn_bnc_ops} shows the best results for fixed operation budgets.
On \emph{usps} and \emph{mnist}, CNNs with ReLU activation require at least $2$--$4\times$ and $4$--$8\times$ as many operations as a BN, respectively, to achieve a better performance.
For the sign activation, an even larger number of operations is required to match the accuracy of the BN.
Moreover, CNNs require many operations to achieve their full potential.
On \emph{usps}, CNNs require at least $64\times$ the operations, and on \emph{mnist}, they even require $256\times$ the operations to achieve their best performance.

\subsection{Model-size-aware TAN structure learning} \label{eq:tan_structure_learning}
\begin{figure*}[!ht]
\centering
\subfloat[letter]{\includegraphics[scale=0.52]{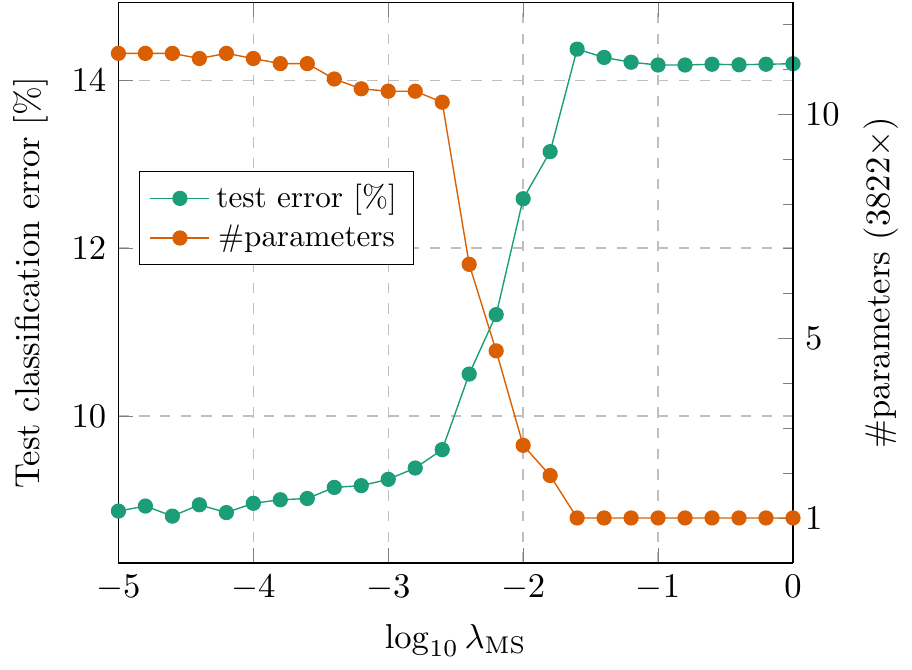}\label{fig:letter_structure_tradeoff_allParams}}
\hfil
\subfloat[satimage]{\includegraphics[scale=0.52]{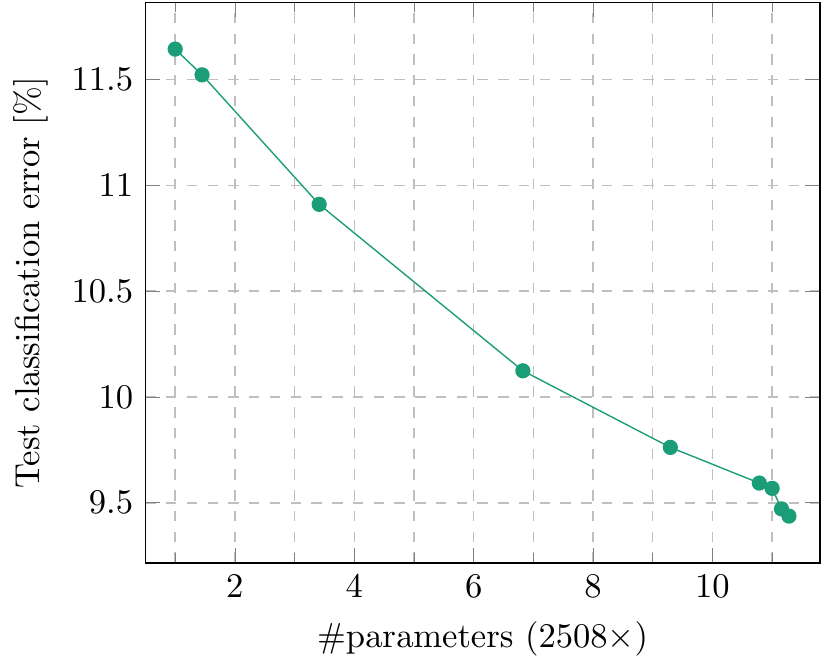}\label{fig:satimage_structure_pareto_allParams}}
\hfil
\subfloat[usps]{\includegraphics[scale=0.52]{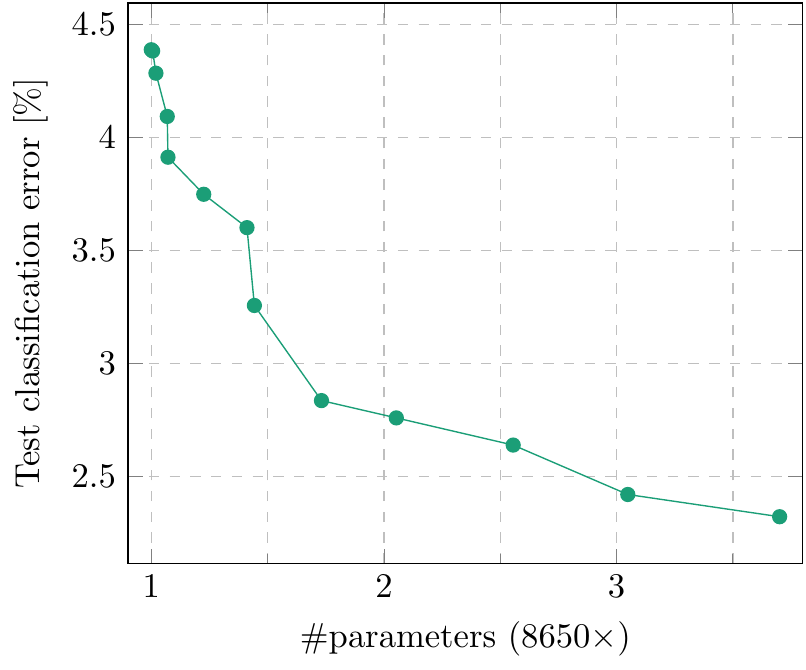}\label{fig:usps_structure_pareto_allParams}}
\hfil
\subfloat[mnist]{\includegraphics[scale=0.52]{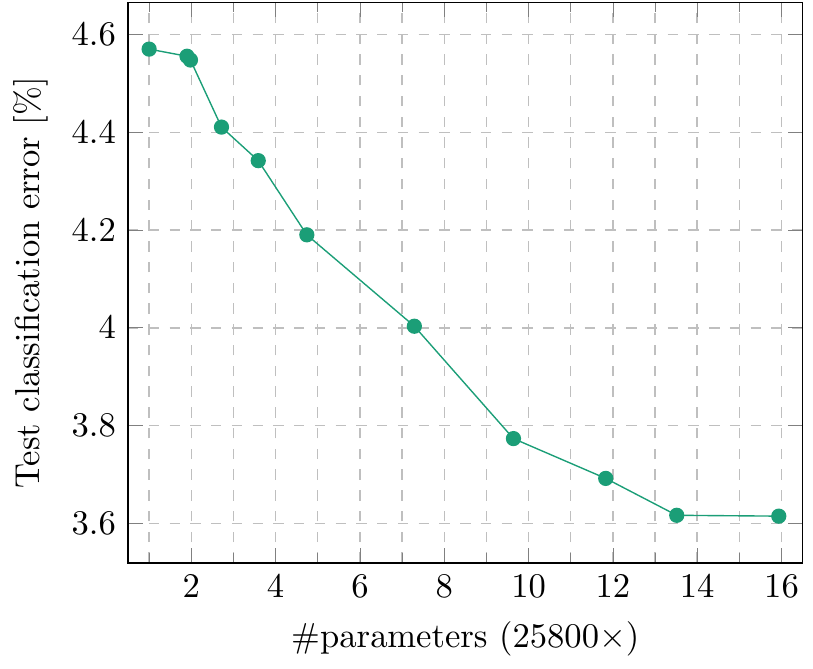}\label{fig:mnist_structure_pareto_allParams}}
\caption{%
Model-size-aware TAN structure learning for BN classifiers.
The number of parameters are shown as multiples of those required by the na\"ive Bayes structure.
\protect\subref{fig:letter_structure_tradeoff_allParams} Test classification error [\%] (left y-axis) and number of parameters (right y-axis) over model size trade-off parameter $\structuretradeoff$ on \emph{letter}.
\protect\subref{fig:satimage_structure_pareto_allParams}--\protect\subref{fig:mnist_structure_pareto_allParams} Pareto optimal models with respect to model size and test classification error obtained by evaluating several $\structuretradeoff$ on the remaining datasets.
\vspace{-10mm}
}
\label{fig:structure_learning}
\end{figure*}

\begin{figure*}[!ht]
\centering
\subfloat[letter]{\includegraphics[scale=0.52]{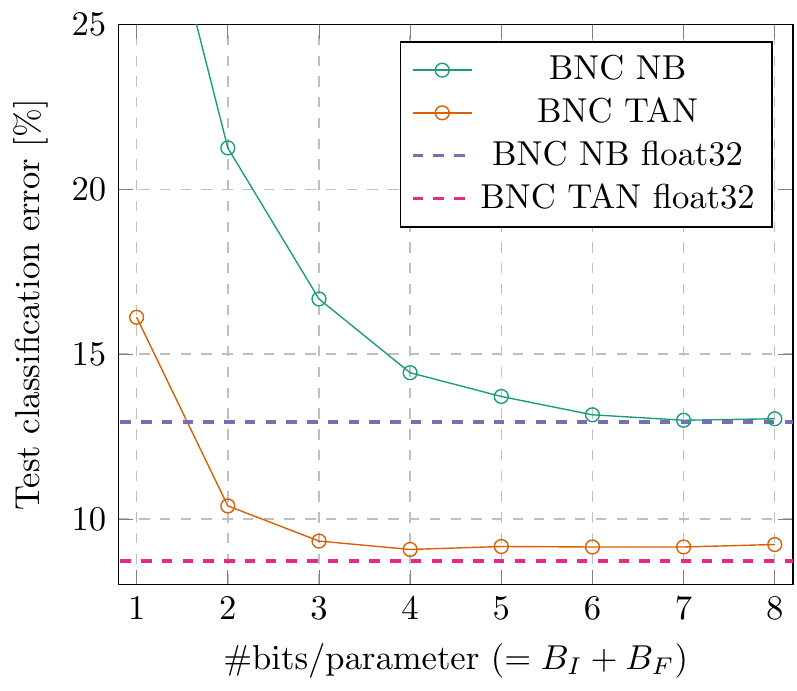}\label{fig:letter_nb_vs_tan_equal_bits}}
\hfil
\subfloat[satimage]{\includegraphics[scale=0.52]{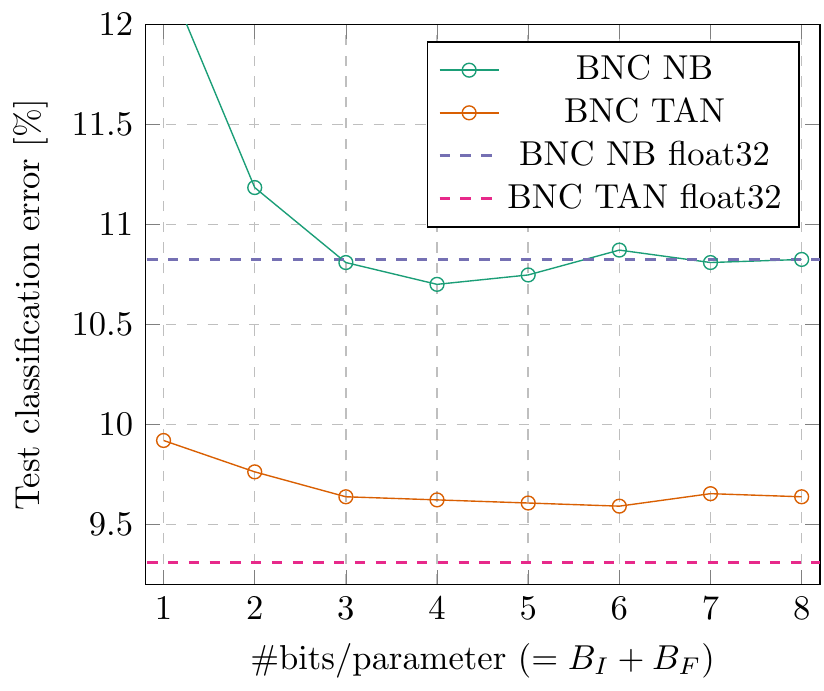}\label{fig:satimage_nb_vs_tan_equal_bits}}
\hfil
\subfloat[usps]{\includegraphics[scale=0.52]{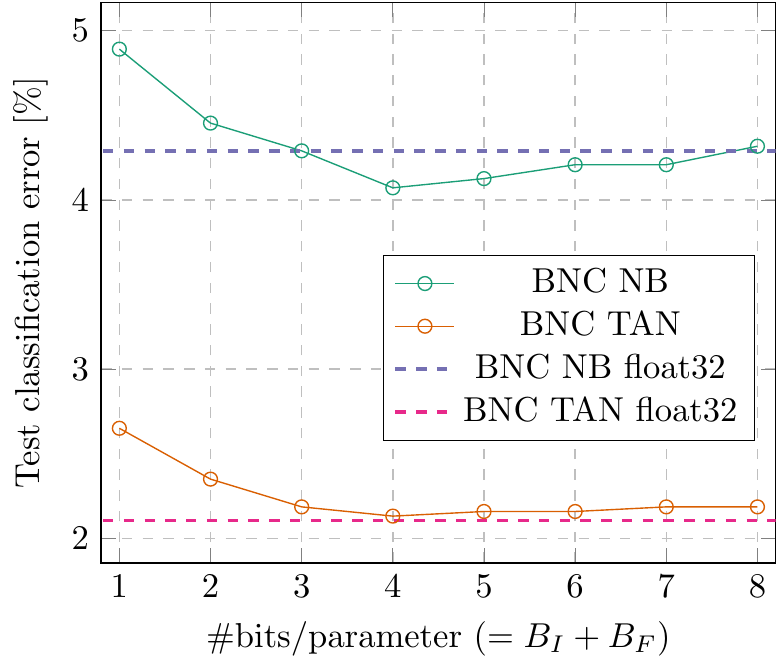}\label{fig:usps_nb_vs_tan_equal_bits}}
\hfil
\subfloat[mnist]{\includegraphics[scale=0.52]{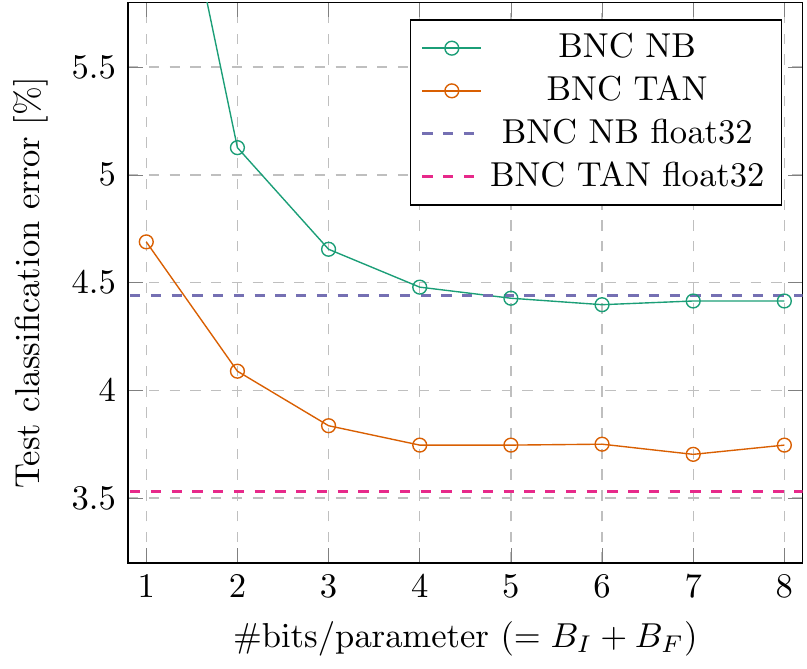}\label{fig:mnist_nb_vs_tan_equal_bits}}
\caption{%
Test classification errors [\%] over numbers of bits per parameter $B_I + B_F$ for quantized BN classifiers (BNC) with na\"ive Bayes (NB) and TAN structure.
The horizontal lines show the respective test errors for float32 parameters.
\vspace{-10mm}
}
\label{fig:compare_bnc_nb_tan_equal_bits}
\end{figure*}

\begin{figure*}[!ht]
\centering
\subfloat[letter]{\includegraphics[scale=0.52]{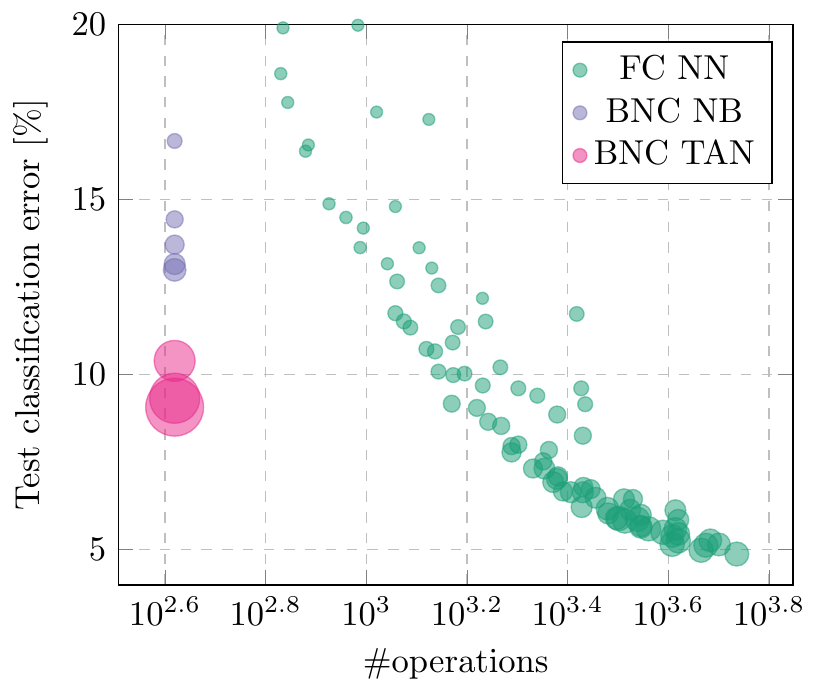}\label{fig:letter_compare_nn_bnc_pareto_ops_err_mem}}
\hfil
\subfloat[usps]{\includegraphics[scale=0.52]{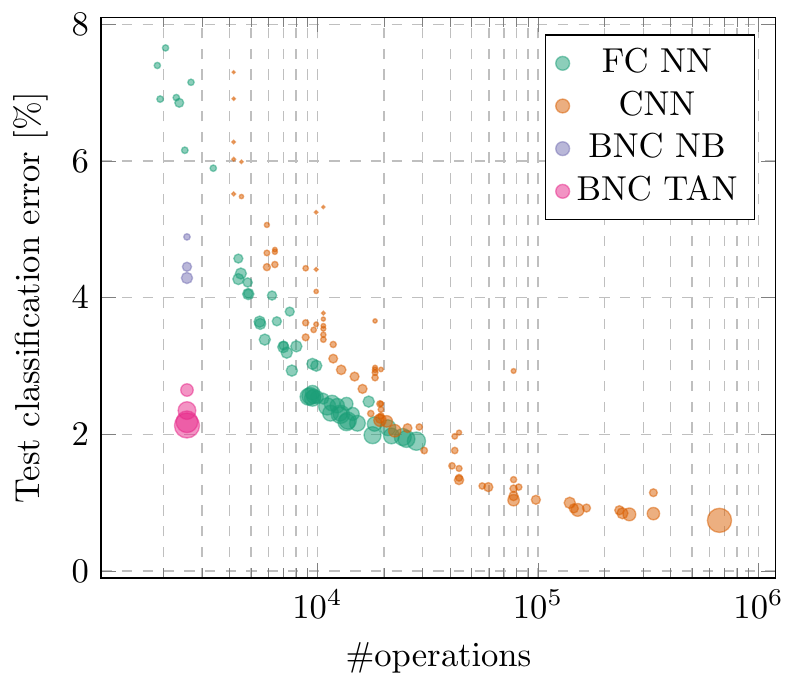}\label{fig:usps_compare_nn_bnc_pareto_ops_err_mem}}
\hfil
\subfloat[satimage]{\includegraphics[scale=0.52]{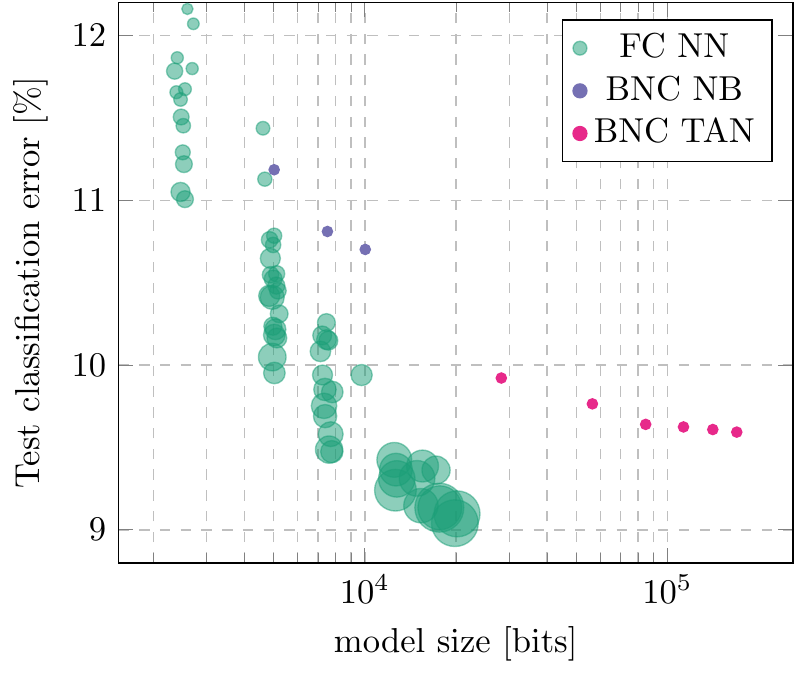}\label{fig:satimage_compare_nn_bnc_pareto_mem_err_ops}}
\hfil
\subfloat[mnist]{\includegraphics[scale=0.52]{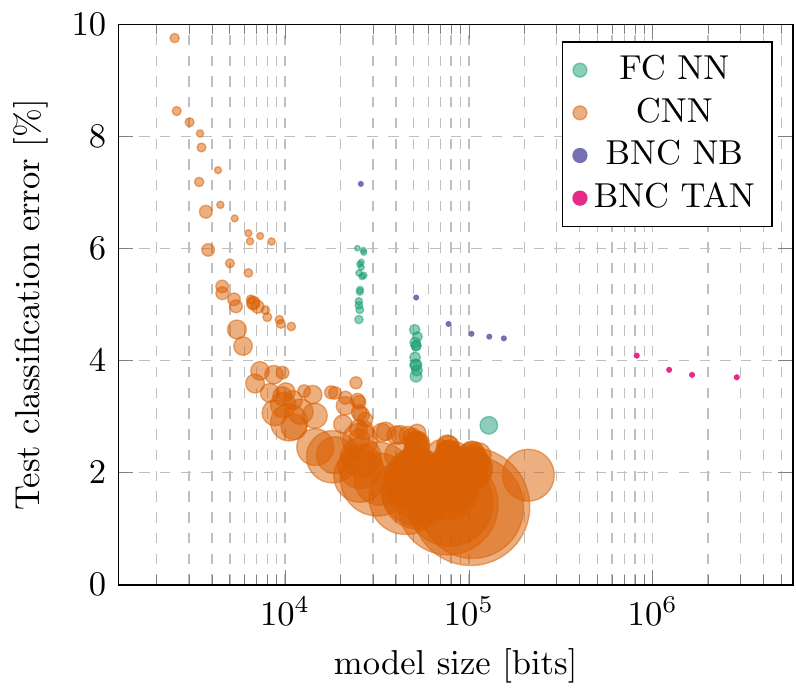}\label{fig:mnist_compare_nn_bnc_pareto_mem_err_ops}}
\caption{%
Comparison of BN classifiers (BNCs) and DNNs.
Each disk corresponds to a Pareto optimal model with respect to test error, number of operations, and parameter memory.
\protect\subref{fig:letter_compare_nn_bnc_pareto_ops_err_mem}, \protect\subref{fig:usps_compare_nn_bnc_pareto_ops_err_mem}: Test classification errors [\%] over number of operations required to compute predictions.
The area of the disks is proportional to the model size in bits.
\protect\subref{fig:satimage_compare_nn_bnc_pareto_mem_err_ops}, \protect\subref{fig:mnist_compare_nn_bnc_pareto_mem_err_ops}: Pareto optimal models with model size on the x-axis and number of operations encoded as the area of the disks.
\vspace{-3.5mm}
}
\label{fig:compare_nn_bnc_pareto}
\end{figure*}

Next, we perform TAN structure learning according to the method described in Section~\ref{sec:structure_learning}.
Note that this experiment uses float32 parameters, i.e., no quantization is involved.
For each $X_i$, we consider fixed randomly selected subsets of possible parents $X_j$ of maximum size $8$ which is called \emph{TAN Subset} in \cite{Roth2020b}.
The hyperparameters of $\loss_{\mathrm{HYB}}$, the feature ordering, and the subsets of possible parents are obtained from the best \emph{TAN Subset} experiment of \cite{Roth2020b}.
We evaluated several trade-off parameters $\structuretradeoff$ to obtain different model sizes and accuracies.

\figurename~\ref{fig:letter_structure_tradeoff_allParams} shows how the accuracy and the test errors vary with $\structuretradeoff$ on \emph{letter} (results are qualitatively similar on the other datasets).
For small $\structuretradeoff$, we obtain an unconstrained TAN structure, whereas for large $\structuretradeoff$, we recover the na\"ive Bayes structure.
For intermediate $\structuretradeoff$, we observe increasing test errors and decreasing model sizes as $\structuretradeoff$ increases.

\figurename~\ref{fig:satimage_structure_pareto_allParams}--\ref{fig:mnist_structure_pareto_allParams} show the Pareto frontier with respect to model size and accuracy by varying $\structuretradeoff$ on \emph{satimage}, \emph{usps}, and \emph{mnist}, respectively.
We note that the leftmost point in each figure corresponds to the na\"ive Bayes model discovered for large $\structuretradeoff$.
Especially on \emph{usps}, a negligible increase in model size is sufficient to achieve substantial gains in accuracy compared to the na\"ive Bayes structure.
Since the CPTs grow by a factor of the number of possible parent values, the granularity of the achievable trade-offs is dataset dependent.
On \emph{usps}, the average number of values per feature is relatively low (i.e., $3.4$), and we can trade off smoothly between model size and accuracy (note the x-axis scale).
On \emph{letter}, \emph{satimage}, and \emph{mnist}, the corresponding numbers of values per feature are larger (i.e., $9.1$, $11.5$ and $13.2$, respectively).

\subsection{Quantization for BN classifiers} \label{sec:quantization_bnc}
\figurename~\ref{fig:compare_bnc_nb_tan_equal_bits} shows test errors of quantized BN classifiers with na\"ive Bayes and TAN structures.
For each dataset, we used a fixed TAN structure obtained from the best \emph{TAN Subset} experiment of \cite{Roth2020b}.
Consequently, the test errors achieved by the float32 TAN BN classifiers are rather optimistic, which explains the consistent performance gap to the quantized models on \emph{satimage} and \emph{mnist}.

Our quantization approach allows us to effectively trade off between accuracy and model size for both BN architectures.
The number of bits at which the test error saturates depends, in addition to the dataset, also on the architecture.
The na\"ive Bayes model is already prone to underfitting such that it suffers more severely than the more expressive TAN structure when using only one or two bits.

\subsection{Comparing DNNs and BN classifiers} \label{sec:comparison_dnns_bns}
Finally, we contrast DNNs and BN classifiers with respect to (i) number of bits to store the parameters, (ii) number of operations, and (iii) test error.
\figurename~\ref{fig:compare_nn_bnc_pareto} shows Pareto optimal models with respect to these three dimensions, i.e., we cannot improve on these models in one dimension without degrading some other dimension.
The models were obtained from the experiments in Sections \ref{sec:experiment_fixed_parameter_memory}, \ref{sec:experiment_fixed_num_operations}, and \ref{sec:quantization_bnc}.
We do not report results for DNNs operating on one-hot encoded inputs.

BN classifiers require very few operations and achieve a moderate test error.
Among BNs, we can improve the performance by selecting a TAN structure instead of a na\"ive Bayes structure, but this typically incurs a considerable memory overhead.
For instance, on \emph{mnist} where the average number of values per feature is relatively high (13.2), it is questionable whether the performance gain can be justified, considering that the memory increases by an order of magnitude.

At the same time, DNNs outperform BNs on every dataset in terms of accuracy, but they require substantially more operations to do so.
Fully connected DNNs allow for a fine-grained trade-off between accuracy, memory, and operations due to their flexible structure.
However, as discussed in Section~\ref{sec:experiment_fixed_parameter_memory}, the memory efficiency of fully connected DNNs can partly be explained by the fact that they consider the inputs as real-valued quantities.
Interestingly, by introducing a \emph{bottleneck} layer exhibiting fewer units than there are output classes, DNNs might even require fewer operations than BNs.
This can be seen, for instance, in \figurename~\ref{fig:usps_compare_nn_bnc_pareto_ops_err_mem} on \emph{usps}.
However, the accuracy degradation in this case is also quite substantial.

Once again, we can see that CNNs are extremely memory efficient, but they require many operations.
For instance, on \emph{mnist}, CNNs require up to three orders of magnitude more operations than BNs to achieve their best accuracy.

\section{Conclusion} \label{sec:conclusion}
We have introduced quantization-aware training for BN classifiers based on the STE which has recently become popular for quantization in DNNs.
We highlighted the effectiveness of our approach in extensive experiments and improved over a specialized branch-and-bound algorithm for learning discrete BN classifiers by a large margin.
Moreover, we contrasted quantized BN classifiers with quantized DNNs and identified regimes of model size, number of operations, and test error in which each model class performs best.
In particular, BN classifiers require few operations and achieve decent accuracy, CNNs are memory efficient and achieve the lowest error, and fully connected DNNs provide flexible trade-offs.
Our results show that quantized DNNs perform well in small-scale scenarios which are hardly investigated in the literature.
Furthermore, we extended previous work on TAN structure learning by incorporating a model size penalty which allows us to effectively trade off between test error and model size.

We have pointed out similarities between BN classifiers and DNNs, which motivates the transfer of well-established techniques from DNNs to BNs.
We believe that several other techniques from the deep learning community, e.g., those discussed in \cite{Roth2020a}, can be successfully transferred to BNs.

\section*{Acknowledgment}
This work was supported by the Austrian Science Fund (FWF) under the project number I2706-N31.




%

\bibliographystyle{IEEEtran}

\bibliography{references}

%

\end{document}